%% file: nl2logic.tex
\pdfoutput=1

\documentclass[11pt]{article}

\usepackage{ACL2024}
\usepackage{times}
\usepackage{latexsym}

\usepackage[T1]{fontenc}

\usepackage[utf8]{inputenc}

\usepackage{microtype}

\usepackage{inconsolata}

\usepackage{amsmath}
\usepackage{amssymb}

\usepackage{graphicx}

\usepackage{algorithm}
\usepackage{algpseudocode}

\usepackage{multirow}
\usepackage{multicol}
\usepackage{tabularx}
\usepackage{mdframed}
\usepackage{svg}
\usepackage[most]{tcolorbox}
\svgsetup{inkscapelatex=false}
\usepackage{multirow}

\NewDocumentCommand{\heng}
{ mO{} }{\textcolor{red}{\textsuperscript{\textit{Heng}}\textsf{\textbf{\small[#1]}}}}

\title{Entailment-Preserving First-order Logic Representations in Natural Language Entailment}

\author{Jinu Lee, Qi Liu, Runzhi Ma, Vincent Han, Ziqi Wang, Heng Ji, Julia Hockenmaier \\
  University of Illinois Urbana-Champaign\\
  \texttt{\{jinulee2, ziqiw9, hengji, juliahmr\}@illinois.edu}
}

\newcommand{\nltofol}{NL$\rightarrow$FOL}

\begin{document}
\maketitle
\begin{abstract}
\input{content/0_abstract}
\end{abstract}


\input{content/1_introduction}

\input{content/2_related_works}

\input{content/3_methods}

\input{content/4_experiment_settings}

\input{content/5_results}

\input{content/6_conclusion}

\bibliography{anthology,custom}
\bibliographystyle{acl_natbib}

\clearpage

\appendix

\input{content/appendix_1_epr}
\input{content/appendix_2_hyperparam}
\input{content/appendix_3_prompts}
\input{content/appendix_4_examples}
\input{content/appendix_5_license}

\end{document}

%% file: content/0_abstract.tex
First-order logic (FOL) can represent the logical entailment semantics of natural language (NL) sentences, but determining \textit{natural language entailment} using FOL remains a challenge. To address this, we propose the Entailment-Preserving FOL representations (EPF) task and introduce reference-free evaluation metrics for EPF, the Entailment-Preserving Rate (EPR) family. In EPF, one should generate FOL representations from multi-premise natural language entailment data (\textit{e.g.} EntailmentBank) so that the automatic prover's result preserves the entailment labels. Experiments show that existing methods for NL-to-FOL translation struggle in EPF. To this extent, we propose a training method specialized for the task, \textit{iterative learning-to-rank}, which directly optimizes the model’s EPR score through a novel scoring function and a learning-to-rank objective. Our method achieves a 1.8–2.7\% improvement in EPR and a 17.4–20.6\% increase in EPR@16 compared to diverse baselines in three datasets.
Further analyses reveal that iterative learning-to-rank effectively suppresses the arbitrariness of FOL representation by reducing the diversity of predicate signatures, and maintains strong performance across diverse inference types and out-of-domain data.

%% file: content/1_introduction.tex
\section{Introduction}


First-order logic (FOL) expressions are frequently used as a semantic representation of natural language (NL) \citep{bos-nli, folio, malls}. FOL representations are well-suited for expressing the semantics of \textit{logical entailment}, where the hypothesis \textit{necessarily follows} from the lexical meaning of the premises and the logical rules (\textit{e.g.} syllogisms). Furthermore, logical entailment can be easily determined with FOL representations by using the automatic theorem prover.

On the other hand, recognizing textual entailment (RTE) tasks \citep{rte, esnli, entailmentbank} adopt a broader definition of entailment, hereby referred to as \textbf{natural language entailment}. Natural language entailment can be defined as: "$P$ entails $h$ if a human reading $P$ would infer that $h$ \textit{is most likely true}" \citep{rte}, which is a strictly looser condition compared to logical entailment.

\begin{figure}[tp]
    \centering
    \includegraphics[width=\linewidth]{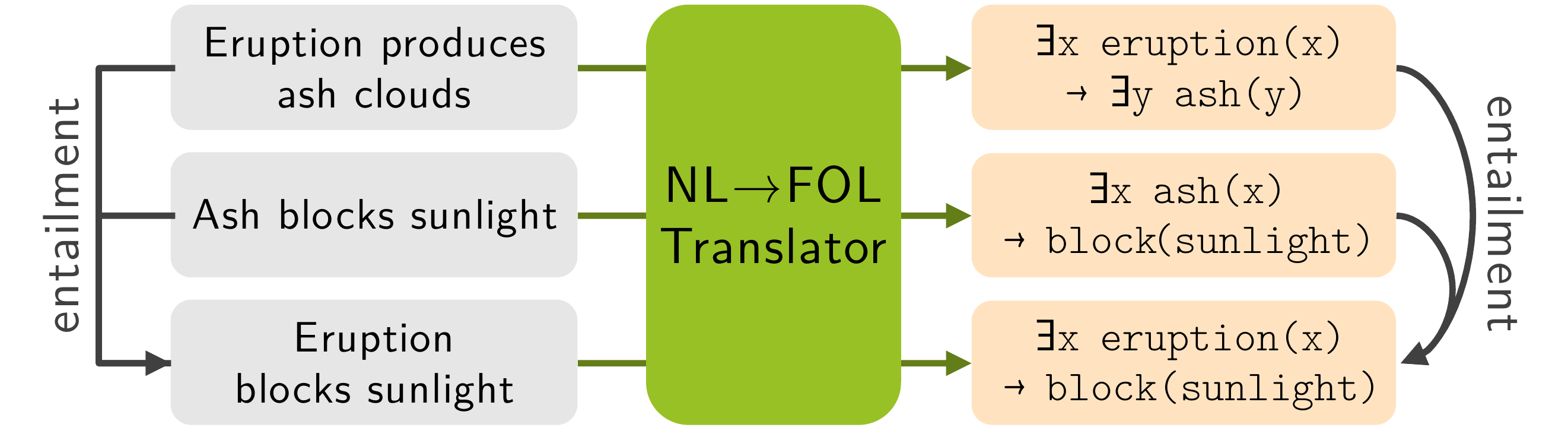}
    \caption{Overview of Entailment-Preserving FOL representations (EPF). When premises entail a hypothesis (gray), the model (green) should produce FOL representations (orange) that preserve the entailment, which can be checked by an automatic theorem prover.}
    \label{fig:entailment-preserving}
\end{figure}

Can we determine natural language entailment using FOL representations?
To date, no prior work on \nltofol\ translation has successfully tackled the question. Classic \nltofol\ parsing approaches that translate syntactic and semantic parses to first-order logic often failed to preserve entailment in single-premise RTE tasks \citep{bos-markert-2005-recognising, bos-nli}. 
Recent methods use large language models (LLMs) to generate FOL representations from NL, and apply an automatic theorem prover to prove or disprove the given hypothesis \citep{linc, logiclm}. While these methods achieve high performance in multi-premise logical entailment tasks \citep{tafjord-etal-2021-proofwriter, folio}, the generalizability of these methods to natural language entailment is not yet tested.

As a systematic approach to this long-standing problem, we formalize the \textbf{Entailment-Preserving FOL representations (EPF)} task. In the EPF task, one must generate FOL representations for each premise and hypothesis in a multi-premise RTE dataset that preserves the entailment label. This task is challenging because as no reference FOL representation is provided, the model can only learn from the execution results of the theorem prover. Along with the task, we present a suite of \textit{reference-free} metrics for the EPF task, namely the \textbf{entailment-preserving rate (EPR)} family.

We empirically show that existing approaches for obtaining FOL representations, including classic meaning representations-based methods and end-to-end generative models, cannot effectively solve the EPF task. To advance the state-of-the-art, we develop a novel \textbf{iterative learning-to-rank} training method. It rewards FOL representations that preserve the entailment and penalizes ones that cannot, pushing the model to produce more entailment-preserving FOL representations. Experiments show that a T5 \citep{t5} model trained using the proposed method significantly outperforms baselines on the EPF task on three multi-premise entailment datasets (EntailmentBank \citep{entailmentbank}, eQASC \citep{eqasc}, and e-SNLI \citep{esnli}). Furthermore, we provide analyses that show the proposed training method can reduce the unwanted arbitrariness in FOL predicates, and generalize to diverse inference types and out-of-domain data.

Our key contributions can be summarized as follows.

\begin{itemize}
    \item We formalize the Entailment-Preserving FOL representations (\textbf{EPF}) task, where one must produce FOL representations that preserve the entailment in multi-premise RTE datasets. We empirically show that the task is challenging for diverse \nltofol\ translator baselines.
    \item We develop a suite of reference-free metrics, Entailment-Preserving Rate (EPR) family, for evaluating EPF.
    \item We propose \textbf{iterative learning-to-rank} training for EPF that significantly outperforms baselines. We perform multiple analyses to show that the method effectively reduces arbitrariness and is robust to various in-domain and out-of-domain data distributions.
\end{itemize}

%% file: content/2_related_works.tex
\section{Related Work}

\subsection{First-order logic for natural entailment}

Since the start of the RTE challenge \citep{rte}, multiple works have attempted using FOL representations to solve natural language entailment. These methods first obtain the syntactic/semantic parse tree and apply a rule-based transformation to get the FOL representation \citep{bos-markert-2005-recognising, bos-nli}. However, it was repeatedly shown that these FOL representations are not empirically effective in solving natural language entailment. For instance, \citet{bos-nli} reported that FOL representations translated from the discourse representation structure (DRS) yield only 1.9\% recall in detecting the entailment in the single-premise RTE benchmark \citep{rte}.

Independently from these works, multi-premise logical entailment benchmarks \citep{tafjord-etal-2021-proofwriter, logicnli, folio} were developed to evaluate the reasoning ability of generative models. These benchmarks adopt the classic 3-way entailment label classification format (\textit{entailment, contradiction, neutral}) of single-premise RTE tasks, in which both the NL sentences and their gold FOL representations point to the same entailment label. 

Recent works have applied LLMs to obtain FOL representations for these multi-premise logical entailment tasks \citep{logiclm, linc, divide-and-translate}, fueled by the code generation ability of LLMs. While they achieve significant performance in synthetic, controlled logical reasoning benchmarks, whether they can generalize to natural entailment has remained unanswered. Furthermore, \citet{linc} observed that LLMs are highly susceptible to \textit{arbitrariness}, as they fail to produce coherent predicate names or numbers of arguments even when generating FOL representations of premises and hypotheses in a single inference.

\subsection{Executable semantic representations}

Apart from FOL, a stream of research focuses on the \textit{executability} of semantic representations. From this perspective, semantic representations are \textit{program codes} that can be executed to solve downstream tasks, such as query intent analysis \citep{spider, dligach-etal-2022-exploring} and question answering \citep{semparse-qa}. The performance of the semantic parser is directly assessed by the accuracy of execution results for the downstream tasks, rather than the similarity between the prediction and the reference parse.

To improve the execution accuracy that is often non-differentiable, reinforcement learning (RL) and its variants have been applied to train neural semantic parsers \citep{cheng-etal-2019-learning, cheng-lapata-2018-weakly}. Using only the input sentence and the desired execution result, these methods learn to maximize the probability of the representations that lead to the correct execution result. However, these approaches are not directly applicable to EPF, as EPF requires taking account of \textit{interactions between premises and hypotheses} during execution (\textit{i.e.} theorem proving) while these methods assume that sentences are isolated.

%% file: content/3_methods.tex
\section{Methods}


\subsection{Entailment Preserving Rate (EPR)}

\begin{figure}[t]
    \centering
    \includegraphics[width=0.85\linewidth]{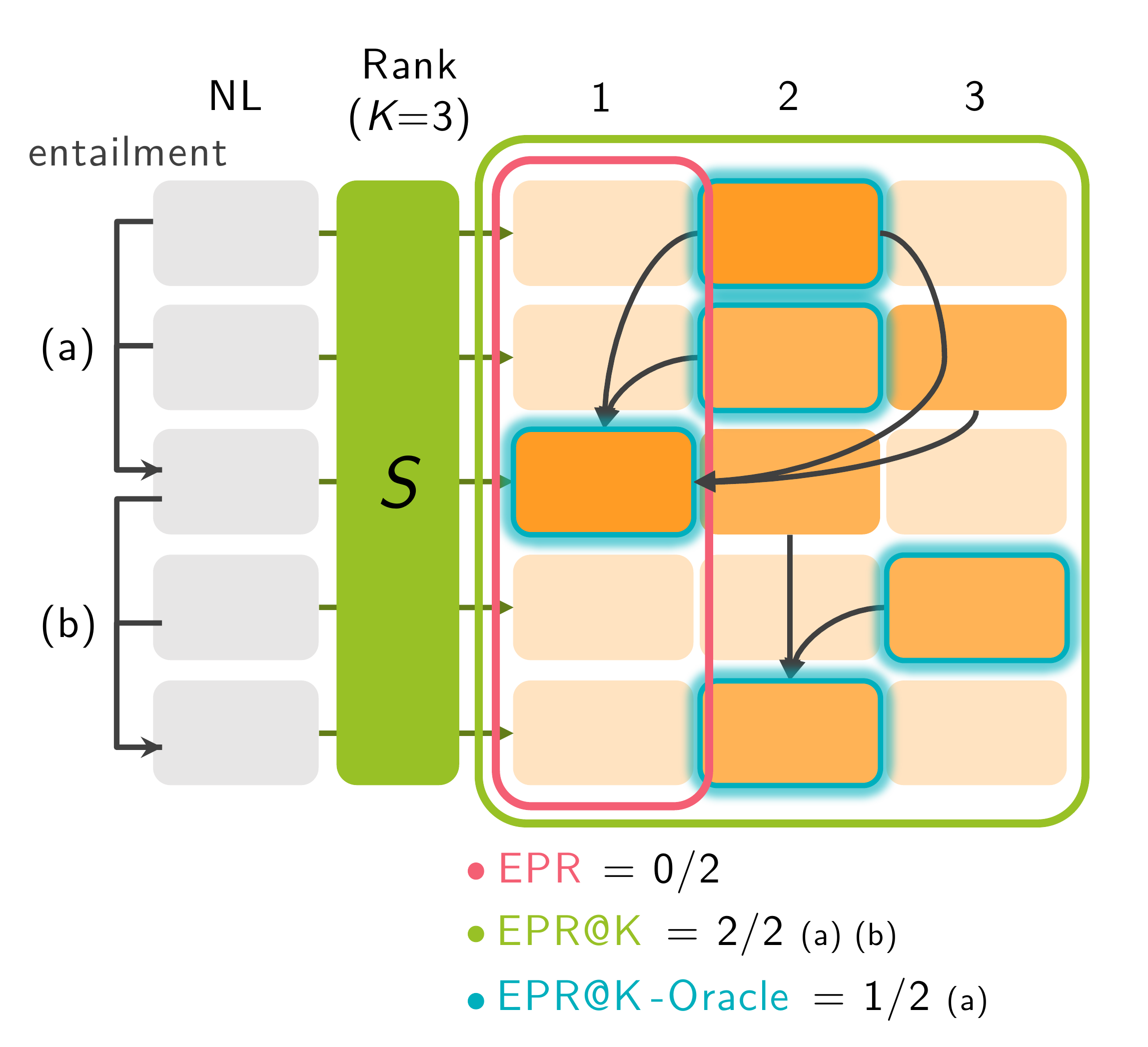}
    \caption{Comparison between EPR, EPR@K, and EPR@K-Oracle. $K=3$ FOL representations (orange) are generated from NL sentences (gray) using the translator $S$, and curved arrows represent entailment-preserving combinations. EPR only uses top 1 predictions for each sentence (red box), where EPR=0/2 because there are no entailment-preserving combinations. EPR@K uses all $K$ predictions (green box) which contain such combinations for both (a) and (b), having a value of 2/2. Finally, EPR@K-Oracle (blue) selects one parse from each sentence that maximizes the global EPR value. In this example, there is no such selection that preserves entailment in both (a) and (b), resulting in EPR@K-Oracle=1/2.}
    \label{fig:metrics}
\end{figure}

Evaluating the quality of FOL representations is challenging because (1) a sentence can have multiple semantically valid FOL representations, and (2) natural language entailment datasets do not usually provide reference FOL representations. In this study, we define \textbf{Entailment-Preserving Rate (EPR)} and its variants, a suite of reference-free metrics that only require entailment labels.

Given an RTE dataset consisting of premises-hypothesis pairs, the EPR of a \nltofol\ translator $S$ is measured by the ratio of the pairs where $S$ can preserve the entailment. Specifically, the translator $S$ translates each premise and hypothesis to FOL representations, obtaining the premise representations $S(P) = S(p_1), ..., S(p_N)$ and the hypothesis representation $S(h)$. Then, an automatic prover is used to determine if $S(h)$ can be proved from $S(P)$\footnote{Contradiction in RTE can be proved by proving $\lnot S(h)$ instead of $S(h)$, therefore it can be treated the same as entailment without loss of generality.}. Finally, a verification step is imposed to filter out spurious entailments caused by contradiction (Appendix \ref{sec:appendix-epr}).

Note that this definition is completely \textit{reference-free} because it does not require a comparison between predicted and gold FOL representations. Furthermore, EPR allows FOL representation to have arbitrary predicate names and logical structures as long as they combine with others to complete a proof, being robust to \textit{arbitrariness}.

We also propose a natural loose extension of EPR to exploit multiple outputs that can be obtained by beam search and sampling. \textbf{EPR@K} allows up to $K$ different parses for each premises and hypothesis. If \textit{any} combination of FOL representations selected from each premise and hypothesis preserves the entailment, it is considered a success.

Finally, \textbf{EPR@K-Oracle} allows only one FOL representation per each premise and hypothesis, similar to EPR. However, instead of selecting the output with the highest model-assigned probability as EPR, outputs are selected from each sentence to maximize the global EPR. This can be viewed as adding an \textit{oracle reranker} to the model. As this constrained optimization problem is NP-complete, we use Answer Set Programming \citep{asp} to get an approximate solution (details in Appendix \ref{sec:appendix-oracle}).

This inequality holds by definition: EPR $=$ EPR@1 $\leq$ EPR@K-Oracle $\leq$ EPR@K. Notably, EPR@K-Oracle serves as an \textit{observable upper bound} for EPR. If a model achieves a specific EPR@K-Oracle score, it indicates the potential existence of a model that achieves an EPR score equal to that value by generating the parses selected by EPR@K-Oracle.





\subsection{Iterative Learning-to-rank}

In this section, we describe the \textbf{iterative learning-to-rank}, a training method specifically designed for the EPF task.

\begin{figure*}[th]
    \centering
    \includegraphics[width=\linewidth]{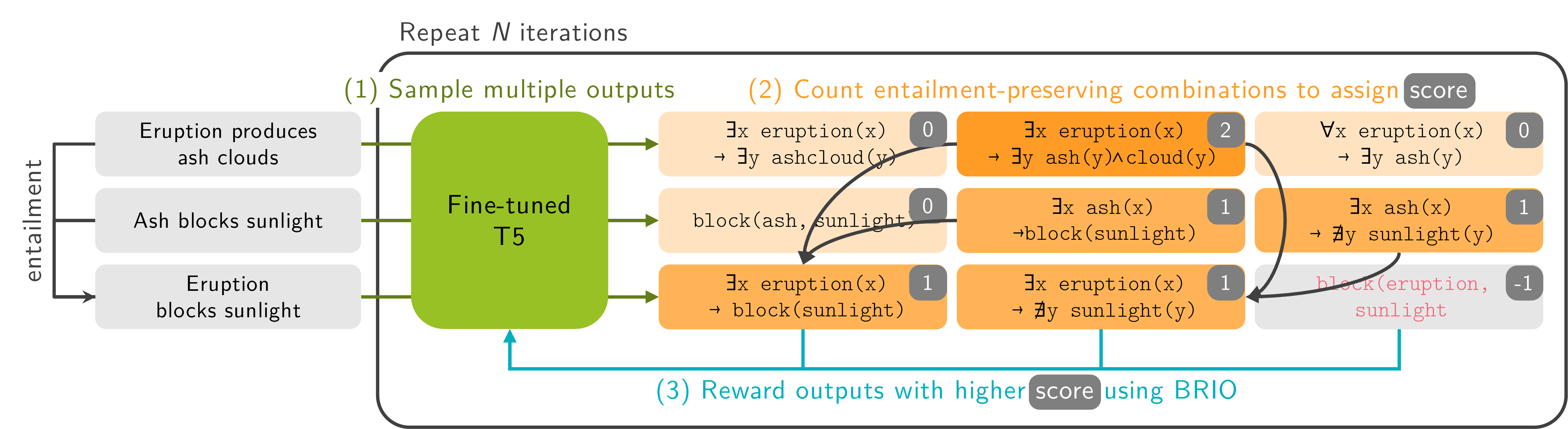}
    \caption{Iterative Learning-to-rank approach to train an entailment-preserving \nltofol\ translator. (1) For each premise and hypothesis, multiple FOL representations are sampled using beam search. (2) An external solver counts all entailment-preserving combinations and assigns scores. (3) Finally, the learning-to-rank objective BRIO is applied to reward the outputs participating in the most entailment-preserving combinations, indirectly increasing the overall EPR. This training loop (1-3) is repeated for multiple iterations to maximize performance.}
    \label{fig:train-loop}
\end{figure*}

\subsubsection{Scoring function}
\label{sec:scoring-func}

The entailment-preserving rate is a dataset-wide metric, but a sentence-to-sentence translation model can only learn from sentence-level rewards. Therefore, we define a sentence-level scoring function such that optimizing it will naturally enhance the global EPR score.

Given a model output $S(p)_{j}$, we define the score as the \textit{number of entailment-preserving combinations} of parses that include $S(p)_{j}$. If the parse contains a syntax error, the score $-1$ is assigned. If a sentence is included in multiple premise-hypothesis pairs, we sum the values obtained from all pairs.

For instance, consider the example in Figure \ref{fig:train-loop}. There are two entailment-preserving combinations annotated with curved arrows. As the second output from \textit{Eruption produces ash clouds.} (darkest orange) is included in both combinations, the score of this output is 2. For other outputs included once (orange), the score 1 is assigned, and outputs that are not included in any combination (lightest orange) get a zero score. Finally, ones that have a syntax error (gray) are assigned a score of -1.




\subsubsection{Learning-to-rank}

If a score is assigned to each output, theoretically any feedback-based learning objective \citep{mrt, ppo, ppr} can be applied to maximize it. In this work, we specifically use BRIO \citep{brio}, a learning-to-rank training objective originally introduced for abstractive summarization models.

Intuitively, in Figure \ref{fig:train-loop}, BRIO tries to increase the average token probability of outputs with higher scores compared to ones with lower scores for each row (same input). Formally, the BRIO training objective $\mathcal{L}_{BRIO}$ is defined as:

\small\[
\mathcal{L}_{BRIO} = \sum_{i}\sum_{j}max(\hat{p}(y_j|x) - \hat{p}(y_i|x) + \Delta(j-i), 0)
\]\normalsize
where $1 \leq i < j \leq K$ denote the indices of outputs $y$ sorted in descending order of the scoring function ($y_1$ having the highest score), $\hat{p}(y|x)$ is the token log-probability normalized by sequence length, and $\Delta$ is the \textit{margin} hyperparameter.

Finally, the plain cross-entropy loss $\mathcal{L}_{CE}$ is added to prevent the model from losing original generation capability, resulting in the final loss function $\mathcal{L} = \mathcal{L}_{CE} + \lambda\mathcal{L}_{BRIO}$ where $\lambda$ is a \textit{mixing rate} hyperparameter. The details of training hyperparameters are included in Appendix \ref{sec:appendix-hyperparam}.

\subsubsection{Iterative training}

Iterative training, which repeats the process of sampling, evaluation, and training, is widely recognized for enhancing performance across various scenarios by enabling the model to deviate further from the original fine-tuned model \citep{iterrpo, multiturn-dpo}.

Initially, a base model $S_{0}$ is obtained by fine-tuning a sequence-to-sequence model on \nltofol\ parallel corpus using only the cross-entropy objective. Then, $S_0$ generates outputs using the training set, which is then evaluated using the scoring function presented in \ref{sec:scoring-func}. After that, a new model $S_{1}$ is trained on the outputs and scores obtained from $S_0$ using the BRIO loss. We repeat this iteration five times, resulting in six different models $S_{t=0..5}$.

%% file: content/4_experiment_settings.tex
\section{Experimental settings}
\subsection{Datasets}
\label{sec:datasets}

\input{table/datasets}

Three representative multi-premise RTE datasets, namely EntailmentBank \citep{entailmentbank}, eQASC \citep{eqasc}, and e-SNLI \citep{esnli}, are used for the experiments. The statistics of each data set are briefly introduced in Table \ref{tab:datasets}.

\textbf{EntailmentBank} provides \textit{entailment trees} with simple scientific facts as nodes. We decompose the trees into depth 1 subtrees which encode a single premises-hypothesis pair.

\textbf{eQASC} provides 2-hop explanations for a given hypothesis derived from QASC \citep{qasc}, a multiple-choice question dataset from the science domain.

\textbf{e-SNLI} extends the single-premise SNLI dataset \citep{snli} by adding \textit{explanations} to the original premise-hypothesis pairs. This can be viewed as the premise and explanation together entailing the hypothesis. Due to limited computation resources, we sample 100k premises-hypothesis pairs from the train set and use the original validation/test set without modification.

\subsection{\nltofol\ translator}

We use a sequence-to-sequence model, T5-base \citep{t5} with 220M parameters, as our \nltofol\ translator backbone. To obtain an initial model $S_0$, we take 34k pairs of natural language sentences and their LLM-generated FOL representation from MALLS \citep{malls},
convert them into the NLTK format \citep{nltk} with a rule-based translator, and train the T5-base model with standard cross-entropy loss. We refer to this model ($S_0$) as \texttt{T5-Iter0}, and models obtained after the $N$-th iteration as \texttt{T5-iter}$N$. Note that the MALLS dataset is not intended to preserve the entailment relationship between sentences, which can be seen in the low EPR score of the initial model \texttt{T5-Iter0} (Table \ref{tab:main-results}).

\subsection{FOL Theorem prover}

For the automatic theorem prover that is used to check entailment, we use Vampire \citep{vampire}, one of the fastest provers currently available. As the generative models are trained on NLTK syntax, the model outputs are translated into Vampire-compatible format (TPTP \citep{tptp}) using a rule-based translator.

\begin{figure*}[t]
    \centering
    \includegraphics[width=\linewidth]{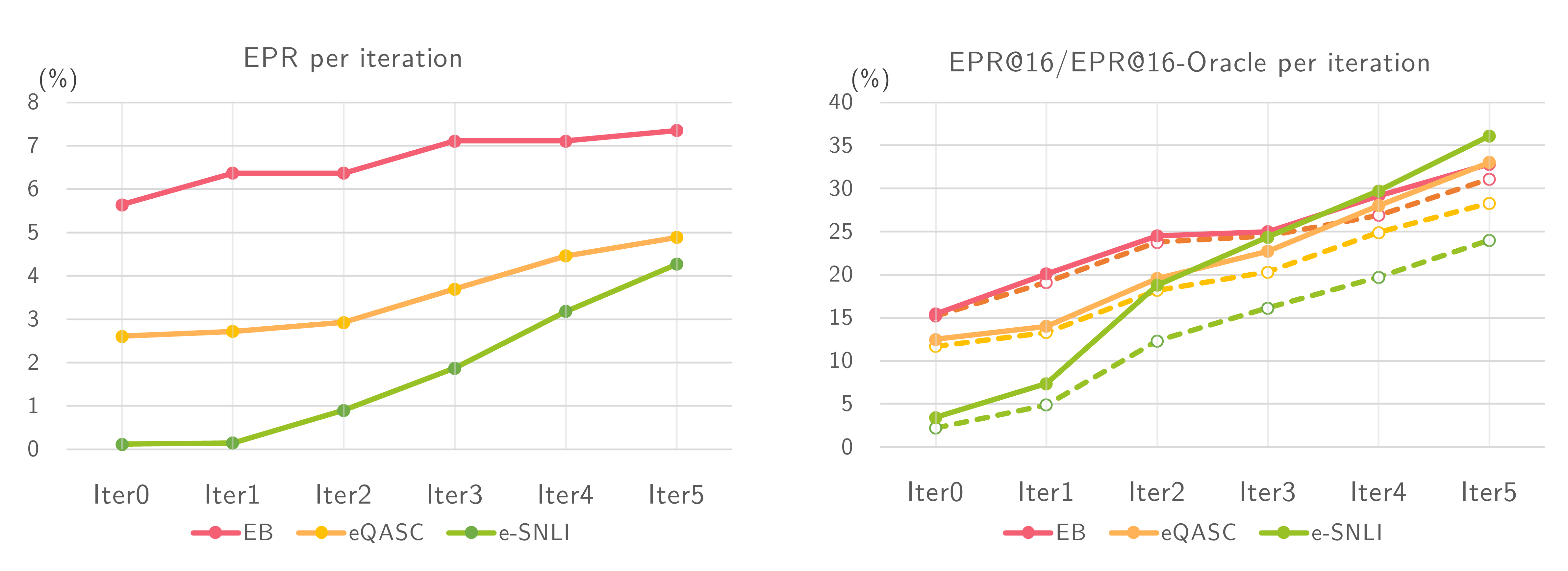}
    \caption{EPR (left), EPR@16 (right-solid), EPR@16-Oracle (right-dotted) per iteration. The continuous growth in all EPR metrics implies that the model extrapolated to unseen premises-conclusion pairs where BRIO loss is 0, demonstrating the strength of the proposed method.}
    \label{fig:main-results}
\end{figure*}

\subsection{Baselines}

As a baseline, we adopt a wide variety of methods that translate natural language to first-order logic.

First, a suite of classic meaning representation-based methods was included as a baseline. \texttt{CCG2Lambda} \citep{ccg2lambda} obtains Combinatory Categorical Grammar (CCG) parse trees using C\&C Parser \citep{cncparser} and converts them to FOL representations via Lambda calculus. \citet{amr2folbos} and \citet{amr2fol} both convert Abstract Meaining Representations (AMR) to FOL. The AMR graph was obtained using AMRBART \citep{amrparser} and translated to FOL representations using the respective implementations of rule-based translators. These methods are only evaluated by EPR and not by EPR@K(-Oracle) as they are designed to produce a single gold FOL parse for each sentence in a deterministic manner. Any errors that occurred during obtaining the FOL representation were removed for the evaluation.

Few-shot and fine-tuned LLMs were also evaluated as a baseline. First, we sample $K=16$ FOL representations using GPT-4o and GPT-4o-mini \citep{gpt4o} with 5 in-context examples temperature 1.0 (prompts shown in Appendix \ref{sec:appendix-prompt}). Also, we evaluate LogicLLaMA \citep{malls}, a LLaMA-7B \citep{llama1} checkpoint directly fine-tuned on the MALLS dataset. $K=16$ FOL representations per each sentence were sampled using temperature 0.1, following the original paper.

%% file: table/datasets.tex
\begin{table}[t]
    \small
    \centering
    \begin{tabular}{l|c|c|c|c}
         \multicolumn{1}{c|}{Dataset} & Train & Valid & Test & Prem. \\
         \hline \hline
         EntailmentBank & 2,486 & 276 & 408 & 2.11 \\
         eQASC & 8,134 & 926 & 920 & 2.00 \\
         e-SNLI & 100,000 & 9,842 & 9,824 & 2.00 \\
    \end{tabular}
    \caption{Dataset statistics. \textbf{Train}, \textbf{Valid}, and \textbf{Test} columns denote the number of premises-hypothesis pairs in each data split. \textbf{Prem.} is the average count of premises included in each premises-hypothesis pair.}
\label{tab:datasets}
\end{table}

%% file: content/5_results.tex

\begin{figure*}[t]
    \centering
    \includegraphics[width=\textwidth]{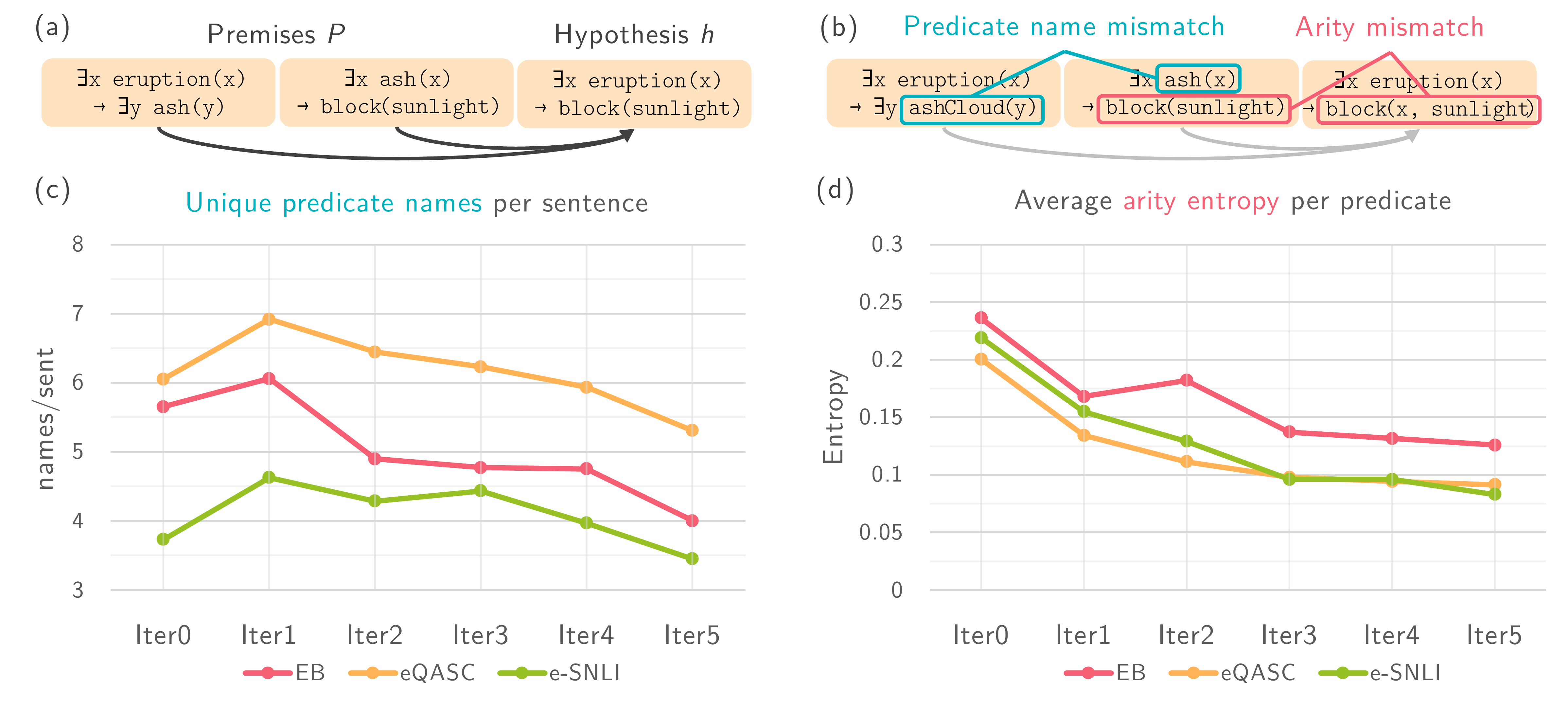}
    \caption{While the FOL premises in \textbf{(a)} can entail the hypothesis (from Figure \ref{fig:entailment-preserving}), \textbf{(b)} cannot due to arbitrariness in predicate name and arity. During the iterative training, \textbf{(c)} the arbitrariness in predicate names decreases after the first iteration, and \textbf{(d)} the arity entropy is significantly reduced. These two results demonstrate that the proposed iterative learning-to-rank method can effectively reduce the arbitrariness of FOL parses.}
    \label{fig:arbitrariness}
\end{figure*}

\section{Results}
\label{sec:result}

\label{sec:epr}

\input{table/epr_results}

The results for EPF on three benchmarks are presented in Table \ref{tab:main-results}.

The results show that classic meaning representations-based methods (\texttt{CCG2Lambda}, \texttt{AMR2FOL}) achieve extremely low EPR in multi-premise RTE datasets, consistently falling short under 0.1\% EPR in EntailmentBank and eQASC. This extends the previous negative results in single-premise RTE datasets \citep{bos-nli, whenlihelps} to multi-premise RTE. On the other hand, end-to-end generative models (\texttt{GPT-4o}, \texttt{LogicLLaMA}, \texttt{T5-Iter0}) also demonstrate low EPR score regardless of the model size or whether it is fine-tuned for \nltofol\ translation or not. Although LLM-based generative methods achieve strong performance in logical entailment tasks \citep{logiclm, linc}, the results show that they do not generalize well to natural language entailment.

As shown in Figure \ref{fig:main-results}, iterative learning-to-rank training can significantly increase the EPR score, resulting in +1.8-4.2p gain in EPR and +22.3-27.8p in EPR@16 after five iterations (\texttt{T5-Iter0}$\rightarrow$\texttt{T5-Iter5}). The BRIO objective provides training signals only for inputs with differing output scores, such as when one output preserves entailment and another does not. The increase of the EPR@K score implies that outputs that previously had zero scores now preserve entailment and have positive scores; indicating that the model \textit{extrapolated} to unseen cases. This demonstrates the effectiveness of the score function defined in Section \ref{sec:scoring-func} and the BRIO learning objective for EPF.

Notably, EPR@16-Oracle scores are significantly higher than EPR and close to the EPR@16 score with only a 1.7p difference in EntailmentBank\footnote{The EPR@16-Oracle scores are a conservative approximation of the \textit{true} EPR@16-Oracle score due to NP-complete evaluation, which favors EntailmentBank that has the smallest test set.}, even though the EPR@16-Oracle score only uses a single FOL parse for each sentence. This implies that the gap between the current state-of-the-art EPR score and the observable upper bound is large, leaving room for future improvement.

Examples of FOL representations sampled from \texttt{T5-Iter0} and \texttt{T5-Iter5} can be found in Appendix \ref{sec:appendix-examples}.

\section{Analysis}
\label{sec:analysis}


\subsection{Arbitrariness}
\label{sec:arbitrariness}

Arbitrariness can be defined as assigning inconsistent predicate signatures (name and arity (number of arguments)) for synonymous concepts. Arbitrariness is critical for generative \nltofol\ translators in the EPF task. For instance, FOL representations in Figure \ref{fig:arbitrariness}(b) cannot entail the hypothesis because predicate names (\texttt{ash}$\leftrightarrow$\texttt{ashCloud}) and arities (\texttt{block} having 1 and 2 arguments) do not match, even though the FOL representations are semantically plausible.
We show that the proposed method effectively reduces arbitrariness in both predicate names and their arity, which contributes to the overall EPR gain.

\subsubsection{Unique predicate names per sentence}

Unique predicate names per sentence can be measured by counting all predicate names and dividing by the number of NL sentences. If the number of unique predicates decreases, it implies that synonymous concepts are mapped to fewer predicates. For instance, unique predicate names per sentence are 1 in Figure \ref{fig:arbitrariness}(a), but 1.33 in Figure \ref{fig:arbitrariness}(b) due to \texttt{ashCloud} and \texttt{ash} being separated.


The results (Figure \ref{fig:arbitrariness}(b)) show that after the first iteration (\texttt{Iter1}), the number of unique predicate names constantly decreases in all datasets, indicating reduced arbitrariness.

\subsubsection{Arity entropy}
End-to-end generative models often fail to generate predicates with consistent \textit{arity}. As the same predicates with different arities cannot lead to a successful proof, it is important to suppress such divergence.

To measure such variance, we adopt a new metric, \textit{arity entropy}. For each predicate, the entropy of the probability distribution of a predicate's arities (\textit{ArityEnt}), \textit{i.e.}
\[
\textit{ArityEnt} = -\sum_{a=1}^{\textrm{max}(a)} p(a)\textrm{log}_2 p(a) \
\]
, was measured. Lower \textit{ArityEnt} indicates that the model prefers a single arity for a given predicate throughout the entire dataset. For instance, \texttt{T5-Iter0} generates \texttt{CausesCycles()} predicate with 2 and 3 arguments 10 and 4 times respectively within the EntailmentBank dataset, resulting in \textit{ArityEnt}=0.86. However, \texttt{T5-Iter5} only generates \texttt{CauseCycles()} with 2 arguments, having \textit{ArityEnt}=0.

Figure \ref{fig:arbitrariness}(c) shows that during iterative training, the average \textit{ArityEnt} of predicates decreases on all three datasets. The result also shows that our method can effectively reduce arbitrariness in order to preserve entailment.

\subsection{Inference types}

\input{table/entailmentbank_inference_types}

\begin{figure}[t]
    \centering
    \includegraphics[width=\linewidth]{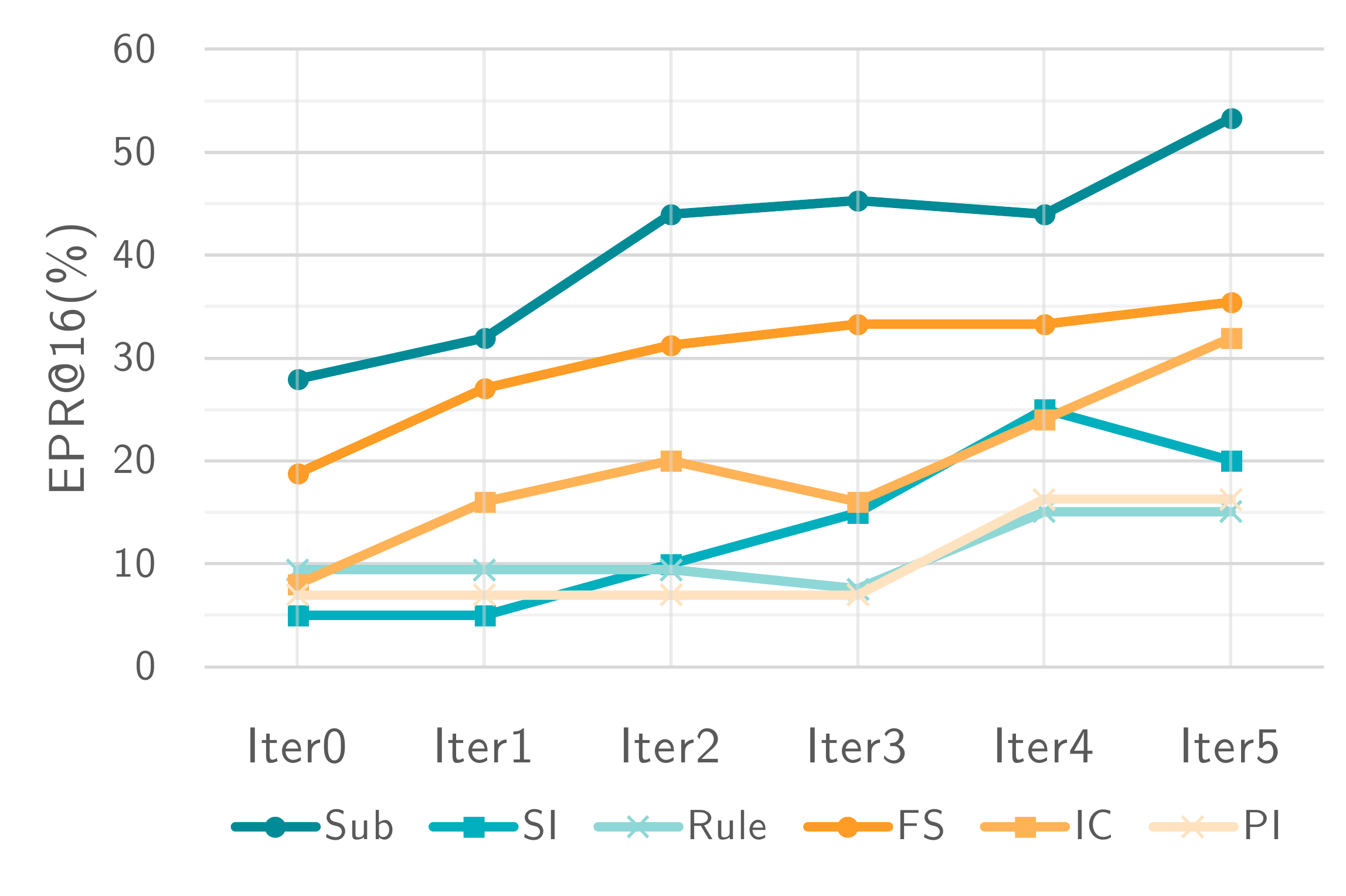}
    \caption{EPR@16 per iteration measured for each inference type in EntailmentBank. Deductive inference (\textit{blue}) and inductive inference (\textit{orange}) both achieve performance gain during the iterative training.}
    \label{fig:reasoning_type}
\end{figure}

To evaluate the robustness of the proposed method against diverse lexical and syntactic patterns, we analyze the EPR in various \textit{inference types}. \citet{entailmentbank} identified three \textit{deductive}, three \textit{inductive} inference types in the EntailmentBank dataset (Table \ref{tab:inference-types}). Deductive inferences are often purely lexical and syntactic as in \textit{An animal is a kind of orgasm. A dog is kind of animal. $\vdash$ A dog is a kind of orgasm.} (Sub), while inductive inferences exhibit non-trivial logical structures, \textit{e.g.} \textit{Hunting is a kind of method for obtaining food. Animals require food for survival. $\vdash$ Some animals must hunt to survive.} (IC) We manually labeled 264 examples from the EntailmentBank test set and measured the EPR@16 score for each reasoning template.

The results (Figure \ref{fig:reasoning_type}) demonstrate that our method consistently improves EPR@16 across all inference types, achieving gains ranging from 5.6p to 25.3p. This shows that our method is robust to the diversity of reasoning patterns (inductive or deductive), and low-resource settings where the proportion of a specific pattern in the training set is as low as 3\%.


\subsection{Out-of-domain generalization}

\begin{figure}[t]
    \centering
    \includegraphics[width=\linewidth]{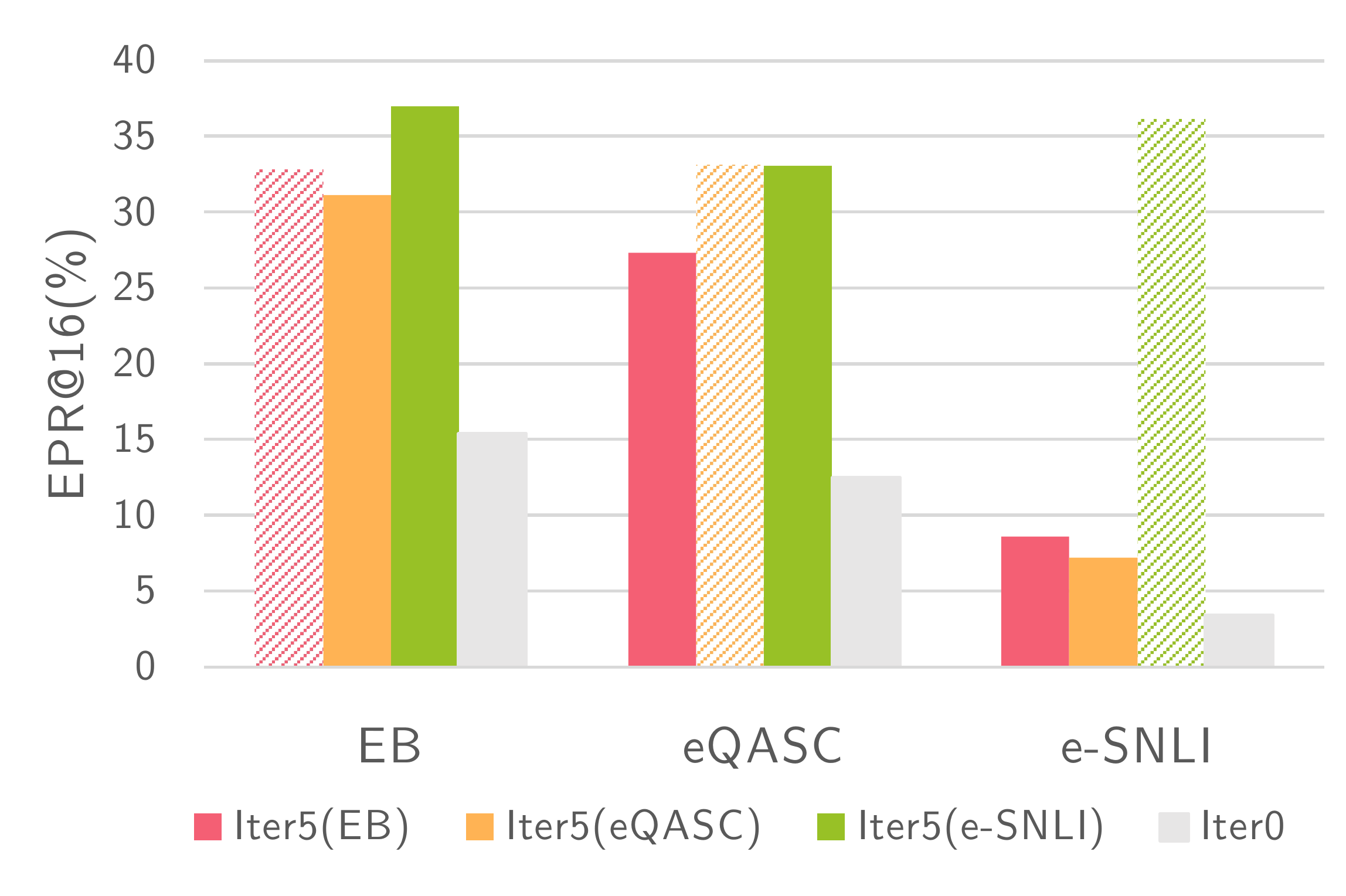}
    \caption{Out-of-domain EPR@16. A model trained on one dataset (\textit{colored}) achieves better performance than the model trained solely on MALLS (\textit{gray}) in other datasets. Hatched items refer to in-domain performance, where the trained and evaluated data are the same.}
    \label{fig:cross_dataset}
\end{figure}

Out-of-domain generalization is crucial for semantic parsers to cover diverse use cases not seen in the training dataset. We evaluate the out-of-domain generalization by evaluating a model trained on one dataset (\textit{e.g.} EntailmentBank) on another dataset (\textit{e.g.} eQASC).

The results are shown in Figure \ref{fig:cross_dataset}. Models trained for five iterations in any dataset consistently outperform \texttt{Iter0} in all out-of-domain settings, which implies that (1) RTE datasets share similar logical structure and (2) the model was able to learn the common structure between multi-premise datasets using the BRIO objective.

The strong out-of-domain generalizability of \texttt{Iter5(e-SNLI)} and relatively low performance on e-SNLI of \texttt{Iter5(EB/eQASC)} demonstrates the effectiveness of e-SNLI for training a \nltofol\ translator. This highlights the importance of wide semantic coverage and large train data in data-driven approaches for the EPF task.

%% file: table/epr_results.tex
\begin{table}[t]
    \small
    \centering
    \begin{tabular}{c|l|ccc}
          Metric & \multicolumn{1}{|c|}{Method} & EB & eQASC & e-SNLI \\
          \hline \hline
          
         \multirow{8}{*}{EPR} &  \texttt{CCG2Lambda} & 0.0 & 0.0 & 0.0 \\
          & \texttt{AMR2FOL(Bos)} & 0.0 & 0.0 & 2.5 \\
          & \texttt{AMR2FOL(Lai)} & 0.0 & 0.0 & 1.6 \\
          & \texttt{GPT-4o-mini} & 3.2 & 2.4 & 0.9 \\
          & \texttt{GPT-4o} & 2.9 & 1.1 & 1.5 \\
          & \texttt{LogicLLaMA} & 5.2 & 2.5 & 0.7 \\
          & \texttt{T5-Iter0} & 5.6 & 2.6 & 0.1 \\
          & \texttt{T5-Iter5} & \textbf{7.4} & \textbf{4.9} & \textbf{4.3} \\
          \hline
         \multirow{5}{*}{EPR@16} & \texttt{GPT-4o-mini} & 10.5 & 7.6 & 8.3 \\
          & \texttt{GPT-4o} & 13.2 & 11.4 & 8.3 \\
          & \texttt{LogicLLaMA} & 5.2 & 2.5 & 0.7 \\
          & \texttt{T5-Iter0} & 15.4 & 12.5 & 3.4 \\
          & \texttt{T5-Iter5} & \textbf{32.8} & \textbf{33.1} & \textbf{36.1} \\
          \hline
         \multirow{5}{*}{\shortstack[c]{EPR@16\\Oracle}} & \texttt{GPT-4o-mini} & 10.5 & 7.4 & 5.6 \\
          & \texttt{GPT-4o} & 13.0 & 10.8 & 5.6 \\
          & \texttt{LogicLLaMA} & 5.2 & 2.5 & 0.7 \\
          & \texttt{T5-Iter0} & 15.2 & 11.7 & 0.1 \\
          & \texttt{T5-Iter5} & \textbf{31.1} & \textbf{28.3} & \textbf{24.0} \\
    \end{tabular}
    \caption{EPR, EPR@16, and EPR@16-Oracle measured on three different datasets (EntailmentBank (EB), eQASC, e-SNLI), single-run.}
    \label{tab:main-results}
\end{table}

%% file: table/entailmentbank_inference_types.tex
\begin{table}[t]
    \small
    \centering
    \begin{tabular}{l|c|c|c}
         \multicolumn{1}{c|}{Type} & D/I & Train Prop. & \# Test  \\
         \hline \hline
         Substitution (Sub) & D & 42\% & 75 \\
         Inference from rule (IR) & D & 33\% & 53 \\
         Further specification (FS) & I & 15\% & 48 \\
         Property inheritance (PI) & I & 4\% & 42 \\
         Infer class (IC) & I & 4\% & 25 \\
         Sequential inference (SI) & D & 3\% & 21 \\
    \end{tabular}
    \caption{Inference types observed from the EntailmentBank dataset, from \citet{entailmentbank}. \textbf{D/I} column indicates if the inference type is deductive (D) or inductive (I). \textbf{Train Prop.} column denotes the proportion of the inference type found in the training set reported by the original paper, and \textbf{\# Test} column is the number of test set examples annotated by the authors.}
\label{tab:inference-types}
\end{table}

%% file: content/6_conclusion.tex
\section{Conclusion}

FOL representations provide an intuitive way to express logical entailment in NL. However, using FOL for checking natural language entailment is a highly complex task, where classic meaning representation-based \nltofol\ parsers and end-to-end generative models both suffer.

In this study, we formalize the Entailment-Preserving FOL representations (EPF) task and reference-free metrics for EPF. Furthermore, we provide an effective method for training an end-to-end generative \nltofol\ translator, \textit{iterative learning-to-rank}, which significantly outperforms baselines. These positive results shed light on a new data-driven approach for understanding natural language entailment with logic, addressing a longstanding challenge in the field.

\section{Limitations}

While our proposed method achieves the state-of-the-art Entailment Preservation Rate (EPR) in Entailment-Preserving FOL representation (EPF) task across multiple datasets and against a broad range of baselines, the gap between the EPR of the proposed approach and the \textit{observed upper bound} performance in EPF (EPR@K-Oracle) still remains. This highlights the significant potential for further advancements in data-driven approaches for \nltofol\ semantic parsing, including the usage of more powerful models and larger training data with broader linguistic coverage.

Moreover, the datasets employed in our study (EntailmentBank, eQASC, e-SNLI) lack linguistically controlled minimal pairs, which are instances designed to highlight subtle but crucial distinctions between sentences. This may cause the \nltofol\ translator to overlook fine-grained differences that are critical for accurately expressing natural language entailment. While previous efforts on linguistically grounded FOL representations \citep{bos-nli, amr2fol} successfully encode human intuition into formal logic, our experimental results reveal that these methods struggle to capture natural entailment. Balancing the empirical strengths of data-driven approaches with the precision of linguistically grounded methods remains an open challenge for future work on the EPF task.

%% file: content/appendix_1_epr.tex
\section{Details on Entailment-Preserving Rate (EPR)}
\label{sec:appendix-epr}

\subsection{Spurious entailment detection}

In first-order logic, a contradictory statement such as $P(a) \land \lnot P(a)$ can derive any FOL formula. Therefore, the notion of entailment-preserving, $FOL(p_1), ..., FOL(p_N) \vdash FOL(h)$, is susceptible to \textit{spurious entailments} where the hypothesis is derived only because there was a contradiction in the premises.

EPR evaluation imposes two verification steps for the external prover's results to filter out these spurious entailments and align the semantics of entailment-preserving to human instincts. First, the hypothesis must not introduce predicates and constants that were not present in the premises. Second, the automatically generated proof of the hypothesis must include all premises. These checks not only prevent trivial contradictions falsely entailing the hypothesis but also more complex cases, such as $P(c), \forall x(\lnot Q(x) \rightarrow (R(x) \land \lnot R(x))) \vdash Q(c)$ where $P(c)$ is not used to prove the conclusion because of an embedded contradiction in $\forall x(\lnot Q(x) \rightarrow (R(x) \land \lnot R(x)))$.

\subsection{EPR@K-Oracle}
\label{sec:appendix-oracle}

EPR@K-Oracle score is calculated by selecting one from the $K$ parses of each sentence that maximizes the overall EPR score. If all sentences are used only in a single premises-hypothesis pair, this score will be identical to EPR@K. However, as some sentences are included in multiple premises-hypothesis pairs, the EPR@K-Oracle score is generally lower than EPR@K.

This problem of finding the EPR@K-Oracle score is an instance of the constraint satisfaction problem that can be reduced to the maximum satisfiability problem, which is NP-complete. The problem can be formulated in three conditions:
\begin{itemize}
    \item For all boolean predicates $\texttt{fol}(i, j)$ that represent $S(p_i)_j$, only one $j$ from each $i$ can be selected. This condition can be expressed as $\forall i\ \exists! j\ \texttt{select}(i, j)$.
    \item If a $(j_1, ..., j_N, j)$ tuple satisfies the $b$-th premises-hypothesis pair in the dataset, \textit{i.e.} $S(p_1)_{j_1}, ..., S(p_N)_{j_N} \vdash S(h)_j$, $\texttt{select}(1, j_1) \land ... \land \texttt{select}(N, j_N) \land \texttt{select}(h, j)$ implies $\texttt{success}(b)$.
    \item The goal is to maximize the number of different \texttt{success(b)} that are true.
\end{itemize}

To solve this constraint optimization problem, we use Answer Set Programming \citep{asp} to formulate the problem and run an ASP solver, Clingo  \citep{clingo}. We impose a 600-second time limit\footnote{Clingo was executed on a single core of AMD EPYC-Milan. Search strategies and heuristics were set as default.} for each constraint satisfaction problem instance, and use the maximum number of $\texttt{success}(b)$ returned from the solver to calculate EPR@K-Oracle.

%% file: content/appendix_2_hyperparam.tex
\section{Training hyperparameters}
\label{sec:appendix-hyperparam}

Table \ref{tab:hyperparams} shows the hyperparameters used for bootstrapping learning-to-rank training proposed in this work. BRIO hyperparameters were obtained from the grid search $\Delta=0, 0.01, 0.1$ and $\lambda=1,10,100$ using the performance after a single iteration (\textit{i.e.} \texttt{T5-Iter1}) in EntailmentBank. After training for 20 epochs with BRIO loss, the checkpoint with the minimum validation loss was selected.

\begin{table}[h]
    \small
    \centering
    \begin{tabular}{c|c}
         \hline
         \multicolumn{2}{l}{\textbf{BRIO Loss}} \\
         \hline
         $\Delta$ & 0.01 \\
         $\lambda$ & 10 \\
         \hline
         \multicolumn{2}{l}{\textbf{Training}} \\
         \hline
         Batch size & 16 \\
         Optimizer & Adam \\
         Learning rate (LR) & 1e-5 \\
         LR Scheduler & Cosine annealing \\
         Gradient clipping & 5.0 \\
         Epoch & 20 \\
    \end{tabular}
    \caption{Training hyperparameters.}
\label{tab:hyperparams}
\end{table}

%% file: content/appendix_3_prompts.tex
\section{Prompts for GPT-4o(-mini)}
\label{sec:appendix-prompt}

Figure \ref{fig:prompts} describes the prompt used for \texttt{GPT-4o} and \texttt{GPT-4o-mini} baseline experiments. For both models, Five few-shot examples were used to perform in-context learning.

\newtcolorbox{fullwidthbox}[1][]{
    width=\textwidth, 
    left=0pt,         
    right=0pt,        
    boxrule=1pt,      
    sharp corners,    
    #1                
}

\begin{figure*}[ht]
\begin{fullwidthbox}[title=System prompt]
\small
You will see a natural language sentence. Translate it into first-order logic (FOL).

You MUST use a common set of predicates to represent the meaning in FOL format.

Below are instructions for the format of FOL logical formulas:

1. \textbf{Variables}: Use lowercase (\texttt{x}, \texttt{y}, etc.) for generic objects.

2. \textbf{Constants}: Use lowercase names (\texttt{john}, \texttt{sun}) for specific entities.

3. \textbf{Predicates}: Represent properties/relations as \texttt{Predicate(arg1, arg2)}, e.g., \texttt{Rises(sun)}, \texttt{Loves(john, mary)}.

4. \textbf{Connectives}:

   - \textbf{Negation (\texttt{-})}: Not, e.g., \texttt{-Rains(x)}
   
   - \textbf{Conjunction (\texttt{\&})}: And, e.g., \texttt{Walks(john) \& Talks(john)}
   
   - \textbf{Disjunction (\texttt{|})}: Or, e.g., \texttt{Walks(john) | Talks(john)}
   
   - \textbf{Implication (\texttt{->})}: If...then, e.g., \texttt{Rains(x) -> Wet(x)}
   
   - \textbf{Biconditional (\texttt{<->})}: If and only if, e.g., \texttt{Rains(x) <-> Wet(x)}
   
5. \textbf{Quantifiers}:

   - \textbf{Universal (\texttt{all})}: For all, e.g., \texttt{all x. (Human(x) >> Mortal(x))}
   
   - \textbf{Existential (\texttt{exists})}: There exists, e.g., `exists x. (Human(x) \& Smart(x))
\end{fullwidthbox}

\begin{fullwidthbox}[title=User prompt (5-shot)]
\small
Few-shot Example 1:

Sentence: All planets orbit a star.

FOL Translation:
\texttt{all x. (Planet(x) -> exists y. (Star(y) \& Orbits(x, y)))}

Few-shot Example 2:

Sentence: Mars is a planet.

FOL Translation:
\texttt{Planet(mars)}

Few-shot Example 3:

Sentence: If a person is a scientist and has access to a laboratory, they can conduct experiments.

FOL Translation:
\texttt{all x. ((Scientist(x) \& HasAccessToLab(x)) -> CanConductExperiments(x))}

Few-shot Example 4:

Sentence: Butterflies are insects.

FOL Translation:
\texttt{all y. (Butterfly(y) -> Insect(y))}

Few-shot Example 5:

Sentence: All insects that have wings can fly.

FOL Translation:
\texttt{all x. (Insect(x) \& HasWings(x) -> CanFly(x))}
\end{fullwidthbox}
\caption{System and user prompts used for GPT-4o and GPT-4o-mini.}
\label{fig:prompts}

\end{figure*}

%% file: content/appendix_4_examples.tex
\input{table/examples}

\section{Examples}
\label{sec:appendix-examples}

Table \ref{tab:examples} shows the examples sampled from three datasets (EntailmentBank, eQASC, and e-SNLI), and shows how the FOL representations changed from \texttt{T5-Iter0} to \texttt{T5-Iter5}.

First, it is noticeable that the entailment-preserving FOL representations have more \textit{atomic} predicate names (\textit{e.g.} \texttt{GlowingBand}), compared to complex predicate names generated by \texttt{T5-Iter0} (\texttt{Appears...NightSky}). This corresponds to the analyses in Section \ref{sec:arbitrariness}, where the bootstrapping reduces the lexical diversity of the predicate names.

Furthermore, example \textbf{e-SNLI\_6529} shows that a new predicate that does not lexically match the original NL premise, $\text{Has}(x, y)$, was introduced by the model. This shows the potential of our method to overcome the \textit{brittleness} of FOL parses, where FOL parses omit underlying commonsense or pragmatic assumptions required for RTE tasks \citep{bos-nli}, by data-driven approaches in the EPF task.

%% file: table/examples.tex
\begin{table*}[ht]
    \small
    \centering
    \renewcommand{\arraystretch}{1.5}
    \begin{tabular}{c|c|l}
    \hline
    \multicolumn{3}{l}{\textbf{EntailmentBank\_26}} \\
    \hline
    NL & $p_1$ & The Milky Way galaxy appears as a glowing band across the night sky. \\
           & $p_2$ & A student observes a glowing band across the night sky. \\
           & $h$ & The student observes the Milky Way galaxy at night. \\
    \hline
    Iter0 & $p_1$ & $\forall x.(\text{MilkyWayGalaxy}(x) \rightarrow \text{AppearsAsGlowingBandAcrossNightSky}(x))$ \\
           & $p_2$ & $\forall x y.((\text{Student}(x) \land \text{Student}(y)) \rightarrow \text{ObservesGlowingBand}(x,y))$ \\
           & $h$ & $\forall x y.((\text{Student}(x) \land \text{MilkyWayGalaxy}(y)) \rightarrow \text{Observes}(x,y))$ \\
    \hline
    Iter5 & $p_1$ & $\forall x.(\text{MilkyWayGalaxy}(x) \rightarrow (\text{GlowingBand}(x) \land \text{NightSky}(x)))$ \\
           & $p_2$ & $\forall x y z.((\text{Student}(x) \land \text{GlowingBand}(y) \land \text{NightSky}(z)) \rightarrow \text{Observes}(x, y))$ \\
           & $h$ & $\forall x y.((\text{Student}(x) \land \text{MilkyWayGalaxy}(y)) \rightarrow  \text{Observes}(x, y))$ \\
    \hline
    \multicolumn{3}{l}{\textbf{eQASC\_536}} \\
    \hline
    NL & $p_1$ & Mammalian teeth are important for digestion. \\
           & $p_2$ & Teeth help aid the digestive system in breaking down food. \\
           & $h$ & Mammalian teeth are important for breaking down food. \\
    \hline
    Iter0 & $p_1$ & $\forall x.(\text{MammalTeeth}(x) \rightarrow \text{ImportantForDigestion}(x))$ \\
           & $p_2$ & $\forall x.(\text{Teeth}(x) \rightarrow \text{AidsDigestiveSystem}(x))$ \\
           & $h$ & $\forall x.(\text{MammalTeeth}(x) \rightarrow \text{BreaksDownFood}(x))$ \\
    \hline
    Iter5 & $p_1$ & $\forall x.(\text{Mammal}(x) \rightarrow (\text{Teeth}(x) \land \text{Digestive}(x)))$ \\
           & $p_2$ & $\forall x.(\text{Teeth}(x) \rightarrow (\text{AidsDigestiveSystem}(x) \land \text{BreaksDownFood}(x)))$ \\
           & $h$ & $\forall x.(\text{Mammal}(x) \rightarrow (\text{Teeth}(x) \land \text{BreaksDownFood}(x)))$ \\
    \hline
    \multicolumn{3}{l}{\textbf{e-SNLI\_6529}} \\
    \hline
    NL & $p_1$ & A woman with red-hair and a leather jacket holds a harp in front of a snowy forest. \\
           & $p_2$ & A harp is an instrument. \\
           & $h$ & The woman is holding an instrument. \\
    \hline
    Iter0 & $p_1$ & $\forall x y.((\text{Female}(x) \land \text{RedHair}(x) \land \text{LeatherJacket}(x) \land \text{SnowyForest}(y)) \rightarrow \text{HoldsHarpInFrontOf}(x,y))$ \\
           & $p_2$ & $\forall x.(\text{Harp}(x) \rightarrow \text{Instrument}(x))$ \\
           & $h$ & $\forall x y.((\text{Female}(x) \land \text{Instrument}(y)) \rightarrow \text{Holds}(x,y))$ \\
    \hline
    Iter5 & $p_1$ & $\exists x y.(\text{Female}(x) \land \text{RedHair}(x) \land \text{LeatherJacket}(x) \land \text{Harp}(y) \land \text{SnowyForest}(z) \land \text{Has}(x,y) \land \text{Holds}(x,y))$ \\
           & $p_2$ & $\forall x.(\text{Harp}(x) \rightarrow \text{Instrument}(x))$ \\
           & $h$ & $\exists x y.(\text{Female}(x) \land \text{Instrument}(y) \land \text{Holds}(x,y))$ \\
    \hline
    \end{tabular}
    \caption{Premises-hypothesis pairs sampled from EntailmentBank, eQASC, and e-SNLI datasets, where the entailment is not preserved in \texttt{T5-Iter0} but preserved in \texttt{T5-Iter5} in EPR@16-Oracle setting.}
    \label{tab:examples}
\end{table*}

%% file: content/appendix_5_license.tex
\section{License}

The datasets (EntailmentBank: Apache 2.0, eQASC: CC-BY-4.0, e-SNLI: MIT, MALLS: CC-BY-NC-4.0), models (T5: Apache 2.0), and theorem prover (Vampire: BSD 3-Clause) can be freely used for non-commercial academic purposes.

%% file: nl2logic.bbl
\begin{thebibliography}{40}
\expandafter\ifx\csname natexlab\endcsname\relax\def\natexlab#1{#1}\fi

\bibitem[{Bai et~al.(2022)Bai, Chen, and Zhang}]{amrparser}
Xuefeng Bai, Yulong Chen, and Yue Zhang. 2022.
\newblock \href {https://aclanthology.org/2022.acl-long.415} {Graph pre-training for {AMR} parsing and generation}.
\newblock In \emph{Proceedings of the 60th Annual Meeting of the Association for Computational Linguistics (Volume 1: Long Papers)}, pages 6001--6015, Dublin, Ireland. Association for Computational Linguistics.

\bibitem[{Bird et~al.(2009)Bird, Klein, and Loper}]{nltk}
Steven Bird, Ewan Klein, and Edward Loper. 2009.
\newblock \emph{Natural language processing with Python: analyzing text with the natural language toolkit}.
\newblock " O'Reilly Media, Inc.".

\bibitem[{Bos(2014)}]{bos-nli}
Johan Bos. 2014.
\newblock \href {https://aclanthology.org/2014.lilt-9.3} {Is there a place for logic in recognizing textual entailment}.
\newblock \emph{Linguistic Issues in Language Technology}, 9.

\bibitem[{Bos(2016)}]{amr2folbos}
Johan Bos. 2016.
\newblock \href {https://doi.org/10.1162/COLI_a_00257} {{S}quib: Expressive power of {A}bstract {M}eaning {R}epresentations}.
\newblock \emph{Computational Linguistics}, 42(3):527--535.

\bibitem[{Bos and Markert(2005)}]{bos-markert-2005-recognising}
Johan Bos and Katja Markert. 2005.
\newblock \href {https://aclanthology.org/H05-1079} {Recognising textual entailment with logical inference}.
\newblock In \emph{Proceedings of Human Language Technology Conference and Conference on Empirical Methods in Natural Language Processing}, pages 628--635, Vancouver, British Columbia, Canada. Association for Computational Linguistics.

\bibitem[{Bos and Markert(2006)}]{whenlihelps}
Johan Bos and Katja Markert. 2006.
\newblock When logical inference helps determining textual entailment (and when it doesn’t).
\newblock In \emph{Proceedings of the second PASCAL RTE challenge}, volume~26.

\bibitem[{Bowman et~al.(2015)Bowman, Angeli, Potts, and Manning}]{snli}
Samuel~R Bowman, Gabor Angeli, Christopher Potts, and Christopher~D Manning. 2015.
\newblock A large annotated corpus for learning natural language inference.
\newblock \emph{arXiv preprint arXiv:1508.05326}.

\bibitem[{Camburu et~al.(2018)Camburu, Rockt\"{a}schel, Lukasiewicz, and Blunsom}]{esnli}
Oana-Maria Camburu, Tim Rockt\"{a}schel, Thomas Lukasiewicz, and Phil Blunsom. 2018.
\newblock \href {https://proceedings.neurips.cc/paper_files/paper/2018/file/4c7a167bb329bd92580a99ce422d6fa6-Paper.pdf} {e-snli: Natural language inference with natural language explanations}.
\newblock In \emph{Advances in Neural Information Processing Systems}, volume~31. Curran Associates, Inc.

\bibitem[{Cheng and Lapata(2018)}]{cheng-lapata-2018-weakly}
Jianpeng Cheng and Mirella Lapata. 2018.
\newblock \href {https://doi.org/10.18653/v1/K18-1035} {Weakly-supervised neural semantic parsing with a generative ranker}.
\newblock In \emph{Proceedings of the 22nd Conference on Computational Natural Language Learning}, pages 356--367, Brussels, Belgium. Association for Computational Linguistics.

\bibitem[{Cheng et~al.(2019)Cheng, Reddy, Saraswat, and Lapata}]{cheng-etal-2019-learning}
Jianpeng Cheng, Siva Reddy, Vijay Saraswat, and Mirella Lapata. 2019.
\newblock \href {https://doi.org/10.1162/coli_a_00342} {Learning an executable neural semantic parser}.
\newblock \emph{Computational Linguistics}, 45(1):59--94.

\bibitem[{Clark and Curran(2007)}]{cncparser}
Stephen Clark and James~R. Curran. 2007.
\newblock \href {https://doi.org/10.1162/coli.2007.33.4.493} {Wide-coverage efficient statistical parsing with {CCG} and log-linear models}.
\newblock \emph{Computational Linguistics}, 33(4):493--552.

\bibitem[{Dagan et~al.(2005)Dagan, Glickman, and Magnini}]{rte}
Ido Dagan, Oren Glickman, and Bernardo Magnini. 2005.
\newblock The pascal recognising textual entailment challenge.
\newblock In \emph{Machine learning challenges workshop}, pages 177--190. Springer.

\bibitem[{Dalvi et~al.(2021)Dalvi, Jansen, Tafjord, Xie, Smith, Pipatanangkura, and Clark}]{entailmentbank}
Bhavana Dalvi, Peter Jansen, Oyvind Tafjord, Zhengnan Xie, Hannah Smith, Leighanna Pipatanangkura, and Peter Clark. 2021.
\newblock Explaining answers with entailment trees.
\newblock In \emph{Proceedings of the 2021 Conference on Empirical Methods in Natural Language Processing}, pages 7358--7370.

\bibitem[{Dligach et~al.(2022)Dligach, Bethard, Miller, and Savova}]{dligach-etal-2022-exploring}
Dmitriy Dligach, Steven Bethard, Timothy Miller, and Guergana Savova. 2022.
\newblock \href {https://doi.org/10.18653/v1/2022.clinicalnlp-1.12} {Exploring text representations for generative temporal relation extraction}.
\newblock In \emph{Proceedings of the 4th Clinical Natural Language Processing Workshop}, pages 109--113, Seattle, WA. Association for Computational Linguistics.

\bibitem[{Gebser et~al.(2019)Gebser, Kaminski, Kaufmann, and Schaub}]{clingo}
Martin Gebser, Roland Kaminski, Benjamin Kaufmann, and Torsten Schaub. 2019.
\newblock Multi-shot asp solving with clingo.
\newblock \emph{Theory and Practice of Logic Programming}, 19(1):27--82.

\bibitem[{Han et~al.(2024)Han, Schoelkopf, Zhao, Qi, Riddell, Zhou, Coady, Peng, Qiao, Benson, Sun, Wardle-Solano, Szabo, Zubova, Burtell, Fan, Liu, Wong, Sailor, Ni, Nan, Kasai, Yu, Zhang, Fabbri, Kryscinski, Yavuz, Liu, Lin, Joty, Zhou, Xiong, Ying, Cohan, and Radev}]{folio}
Simeng Han, Hailey Schoelkopf, Yilun Zhao, Zhenting Qi, Martin Riddell, Wenfei Zhou, James Coady, David Peng, Yujie Qiao, Luke Benson, Lucy Sun, Alex Wardle-Solano, Hannah Szabo, Ekaterina Zubova, Matthew Burtell, Jonathan Fan, Yixin Liu, Brian Wong, Malcolm Sailor, Ansong Ni, Linyong Nan, Jungo Kasai, Tao Yu, Rui Zhang, Alexander~R. Fabbri, Wojciech Kryscinski, Semih Yavuz, Ye~Liu, Xi~Victoria Lin, Shafiq Joty, Yingbo Zhou, Caiming Xiong, Rex Ying, Arman Cohan, and Dragomir Radev. 2024.
\newblock \href {http://arxiv.org/abs/2209.00840} {Folio: Natural language reasoning with first-order logic}.

\bibitem[{Jhamtani and Clark(2020)}]{eqasc}
Harsh Jhamtani and Peter Clark. 2020.
\newblock Learning to explain: Datasets and models for identifying valid reasoning chains in multihop question-answering.
\newblock \emph{arXiv preprint arXiv:2010.03274}.

\bibitem[{Khot et~al.(2020)Khot, Clark, Guerquin, Jansen, and Sabharwal}]{qasc}
Tushar Khot, Peter Clark, Michal Guerquin, Peter Jansen, and Ashish Sabharwal. 2020.
\newblock Qasc: A dataset for question answering via sentence composition.
\newblock In \emph{Proceedings of the AAAI Conference on Artificial Intelligence}, volume~34, pages 8082--8090.

\bibitem[{Kov{\'a}cs and Voronkov(2013)}]{vampire}
Laura Kov{\'a}cs and Andrei Voronkov. 2013.
\newblock First-order theorem proving and vampire.
\newblock In \emph{Computer Aided Verification}, pages 1--35, Berlin, Heidelberg. Springer Berlin Heidelberg.

\bibitem[{Lai et~al.(2020)Lai, Donatelli, and Pustejovsky}]{amr2fol}
Kenneth Lai, Lucia Donatelli, and James Pustejovsky. 2020.
\newblock \href {https://aclanthology.org/2020.dmr-1.1} {A continuation semantics for {A}bstract {M}eaning {R}epresentation}.
\newblock In \emph{Proceedings of the Second International Workshop on Designing Meaning Representations}, pages 1--12, Barcelona Spain (online). Association for Computational Linguistics.

\bibitem[{Lee et~al.(2023)Lee, Lee, and Hwang}]{ppr}
Youngwon Lee, Jinu Lee, and Seung-won Hwang. 2023.
\newblock \href {https://doi.org/10.18653/v1/2023.emnlp-main.371} {Learning to rank generation with pairwise partial rewards}.
\newblock In \emph{Proceedings of the 2023 Conference on Empirical Methods in Natural Language Processing}, pages 6078--6092, Singapore. Association for Computational Linguistics.

\bibitem[{Lifschitz(2019)}]{asp}
Vladimir Lifschitz. 2019.
\newblock \emph{Answer set programming}, volume~3.
\newblock Springer Heidelberg.

\bibitem[{Liu et~al.(2022)Liu, Liu, Radev, and Neubig}]{brio}
Yixin Liu, Pengfei Liu, Dragomir Radev, and Graham Neubig. 2022.
\newblock Brio: Bringing order to abstractive summarization.
\newblock \emph{arXiv preprint arXiv:2203.16804}.

\bibitem[{Mart\'{i}nez-G\'{o}mez et~al.(2016)Mart\'{i}nez-G\'{o}mez, Mineshima, Miyao, and Bekki}]{ccg2lambda}
Pascual Mart\'{i}nez-G\'{o}mez, Koji Mineshima, Yusuke Miyao, and Daisuke Bekki. 2016.
\newblock \href {https://aclweb.org/anthology/P/P16/P16-4015.pdf} {ccg2lambda: A compositional semantics system}.
\newblock In \emph{Proceedings of ACL 2016 System Demonstrations}, pages 85--90, Berlin, Germany. Association for Computational Linguistics.

\bibitem[{Olausson et~al.(2023)Olausson, Gu, Lipkin, Zhang, Solar-Lezama, Tenenbaum, and Levy}]{linc}
Theo~X Olausson, Alex Gu, Benjamin Lipkin, Cedegao~E Zhang, Armando Solar-Lezama, Joshua~B Tenenbaum, and Roger Levy. 2023.
\newblock Linc: A neurosymbolic approach for logical reasoning by combining language models with first-order logic provers.
\newblock \emph{arXiv preprint arXiv:2310.15164}.

\bibitem[{OpenAI(2024)}]{gpt4o}
OpenAI. 2024.
\newblock \href {http://arxiv.org/abs/2410.21276} {Gpt-4o system card}.

\bibitem[{Pan et~al.(2023)Pan, Albalak, Wang, and Wang}]{logiclm}
Liangming Pan, Alon Albalak, Xinyi Wang, and William Wang. 2023.
\newblock \href {https://doi.org/10.18653/v1/2023.findings-emnlp.248} {Logic-{LM}: Empowering large language models with symbolic solvers for faithful logical reasoning}.
\newblock In \emph{Findings of the Association for Computational Linguistics: EMNLP 2023}, pages 3806--3824, Singapore. Association for Computational Linguistics.

\bibitem[{Pang et~al.(2024)Pang, Yuan, Cho, He, Sukhbaatar, and Weston}]{iterrpo}
Richard~Yuanzhe Pang, Weizhe Yuan, Kyunghyun Cho, He~He, Sainbayar Sukhbaatar, and Jason Weston. 2024.
\newblock \href {http://arxiv.org/abs/2404.19733} {Iterative reasoning preference optimization}.

\bibitem[{Raffel et~al.(2020)Raffel, Shazeer, Roberts, Lee, Narang, Matena, Zhou, Li, and Liu}]{t5}
Colin Raffel, Noam Shazeer, Adam Roberts, Katherine Lee, Sharan Narang, Michael Matena, Yanqi Zhou, Wei Li, and Peter~J. Liu. 2020.
\newblock \href {http://jmlr.org/papers/v21/20-074.html} {Exploring the limits of transfer learning with a unified text-to-text transformer}.
\newblock \emph{Journal of Machine Learning Research}, 21(140):1--67.

\bibitem[{Ryu et~al.(2024)Ryu, Kim, Lee, and Yang}]{divide-and-translate}
Hyun Ryu, Gyeongman Kim, Hyemin~S. Lee, and Eunho Yang. 2024.
\newblock \href {http://arxiv.org/abs/2410.08047} {Divide and translate: Compositional first-order logic translation and verification for complex logical reasoning}.

\bibitem[{Schulman et~al.(2017)Schulman, Wolski, Dhariwal, Radford, and Klimov}]{ppo}
John Schulman, Filip Wolski, Prafulla Dhariwal, Alec Radford, and Oleg Klimov. 2017.
\newblock \href {http://arxiv.org/abs/1707.06347} {Proximal policy optimization algorithms}.

\bibitem[{Shen et~al.(2016)Shen, Cheng, He, He, Wu, Sun, and Liu}]{mrt}
Shiqi Shen, Yong Cheng, Zhongjun He, Wei He, Hua Wu, Maosong Sun, and Yang Liu. 2016.
\newblock \href {https://doi.org/10.18653/v1/P16-1159} {Minimum risk training for neural machine translation}.
\newblock In \emph{Proceedings of the 54th Annual Meeting of the Association for Computational Linguistics (Volume 1: Long Papers)}, pages 1683--1692, Berlin, Germany. Association for Computational Linguistics.

\bibitem[{Sutcliffe(2024)}]{tptp}
G.~Sutcliffe. 2024.
\newblock {Stepping Stones in the TPTP World}.
\newblock In \emph{{Proceedings of the 12th International Joint Conference on Automated Reasoning}}, number 14739 in Lecture Notes in Artificial Intelligence, pages 30--50.

\bibitem[{Tafjord et~al.(2021)Tafjord, Dalvi, and Clark}]{tafjord-etal-2021-proofwriter}
Oyvind Tafjord, Bhavana Dalvi, and Peter Clark. 2021.
\newblock \href {https://doi.org/10.18653/v1/2021.findings-acl.317} {{P}roof{W}riter: Generating implications, proofs, and abductive statements over natural language}.
\newblock In \emph{Findings of the Association for Computational Linguistics: ACL-IJCNLP 2021}, pages 3621--3634, Online. Association for Computational Linguistics.

\bibitem[{Tian et~al.(2021)Tian, Li, Chen, Xiao, He, and Jin}]{logicnli}
Jidong Tian, Yitian Li, Wenqing Chen, Liqiang Xiao, Hao He, and Yaohui Jin. 2021.
\newblock \href {https://doi.org/10.18653/v1/2021.emnlp-main.303} {Diagnosing the first-order logical reasoning ability through {L}ogic{NLI}}.
\newblock In \emph{Proceedings of the 2021 Conference on Empirical Methods in Natural Language Processing}, pages 3738--3747, Online and Punta Cana, Dominican Republic. Association for Computational Linguistics.

\bibitem[{Touvron et~al.(2023)Touvron, Lavril, Izacard, Martinet, Lachaux, Lacroix, Rozi{\`e}re, Goyal, Hambro, Azhar et~al.}]{llama1}
Hugo Touvron, Thibaut Lavril, Gautier Izacard, Xavier Martinet, Marie-Anne Lachaux, Timoth{\'e}e Lacroix, Baptiste Rozi{\`e}re, Naman Goyal, Eric Hambro, Faisal Azhar, et~al. 2023.
\newblock Llama: Open and efficient foundation language models.
\newblock \emph{arXiv preprint arXiv:2302.13971}.

\bibitem[{Xiong et~al.(2024)Xiong, Shi, Shen, Rosenberg, Qin, Calandriello, Khalman, Joshi, Piot, Saleh, Jin, Zhang, and Liu}]{multiturn-dpo}
Wei Xiong, Chengshuai Shi, Jiaming Shen, Aviv Rosenberg, Zhen Qin, Daniele Calandriello, Misha Khalman, Rishabh Joshi, Bilal Piot, Mohammad Saleh, Chi Jin, Tong Zhang, and Tianqi Liu. 2024.
\newblock \href {http://arxiv.org/abs/2409.02392} {Building math agents with multi-turn iterative preference learning}.

\bibitem[{Xu et~al.(2014)Xu, Zhang, Feng, and Zhao}]{semparse-qa}
Kun Xu, Sheng Zhang, Yansong Feng, and Dongyan Zhao. 2014.
\newblock Answering natural language questions via phrasal semantic parsing.
\newblock In \emph{CCF International Conference on Natural Language Processing and Chinese Computing}, pages 333--344. Springer.

\bibitem[{Yang et~al.(2023)Yang, Xiong, Payani, Shareghi, and Fekri}]{malls}
Yuan Yang, Siheng Xiong, Ali Payani, Ehsan Shareghi, and Faramarz Fekri. 2023.
\newblock Harnessing the power of large language models for natural language to first-order logic translation.
\newblock \emph{arXiv preprint arXiv:2305.15541}.

\bibitem[{Yu et~al.(2018)Yu, Zhang, Yang, Yasunaga, Wang, Li, Ma, Li, Yao, Roman et~al.}]{spider}
Tao Yu, Rui Zhang, Kai Yang, Michihiro Yasunaga, Dongxu Wang, Zifan Li, James Ma, Irene Li, Qingning Yao, Shanelle Roman, et~al. 2018.
\newblock Spider: A large-scale human-labeled dataset for complex and cross-domain semantic parsing and text-to-sql task.
\newblock \emph{arXiv preprint arXiv:1809.08887}.

\end{thebibliography}
